\pdfoutput=1

\documentclass[11pt]{llncs}
\usepackage[a4paper,twoside=false,top=2.32in,bottom=2.69in,left=2.25in,right=2in]{geometry}
\usepackage{times}
 
\usepackage[colorinlistoftodos]{todonotes}
\usepackage{amssymb}
\usepackage{graphicx}
\usepackage[utf8]{inputenc} 
\usepackage[T1]{fontenc} 
\usepackage{lmodern} 
\usepackage{times}
\usepackage{url}
\usepackage{latexsym}
\usepackage{fixltx2e}
\usepackage{qtree}  
\usepackage{multirow}
\usepackage{tabularx} 
\usepackage{booktabs} 

\usepackage{etoolbox}
\patchcmd{\thebibliography}{\chapter*}{\section*}{}{}
\usepackage[numbers,sort&compress,sectionbib]{natbib}
\usepackage{array}
\usepackage{ragged2e}
\newcolumntype{P}[1]{>{\RaggedRight\hspace{0pt}}p{#1}}

\usepackage{url}
\urldef{\mailsa}\path|{m_laali, kosseim}@encs.concordia.ca|

\usepackage{gb4e} 

\begin{document}

\mainmatter  

\title{Automatic Disambiguation of\\French Discourse Connectives}

\titlerunning{Automatic Disambiguation of French Discourse Connectives}

%
%
\author{Majid Laali \and Leila Kosseim}
\authorrunning{Automatic Disambiguation of French Discourse Connectives}

\institute{Department of Computer Science and Software Engineering, \\
Concordia University, Montreal, Quebec, Canada\\
\mailsa}

%
%

\toctitle{Automatic Disambiguation of French Discourse Connectives}
\tocauthor{Majid Laali, Leila Kosseim}
\maketitle

\begin{abstract}
 Discourse connectives (e.g. \textit{however, because}) are terms that can explicitly convey a discourse relation within a text. While discourse connectives have been shown to be an effective clue to automatically identify discourse relations, they are not always used to convey such relations, thus they should first be disambiguated between \textit{discourse-usage} and \textit{non-discourse-usage}. In this paper, we investigate the applicability of features proposed for the disambiguation of English discourse connectives for French. Our results with the French Discourse Treebank (FDTB) show that syntactic and lexical features developed for English texts are as effective for French and allow the disambiguation of French discourse connectives with an accuracy of 94.2\%.
\end{abstract}

\section{Introduction}
Discourse connectives are often used to signal discourse relations between two textual units. For example, in (\ref{ex:one}) the discourse connective \emph{`ainsi'} conveys a \textsc{result} relation between the two textural units marked in \textit{italic} and \textbf{bold}.

\begin{exe}
	\ex \label{ex:one} \textit{L'élan réformateur, lancé depuis Moscou en 1985, revient, tel un boomerang, vers l'URSS}. \textbf{La Lituanie, la Lettonie et l'Estonie s'ouvrent} \underline{ainsi} \textbf{au multipartisme}.\footnote{\label{note1} All examples are taken from the French Discourse Treebank \cite{danlos15}. } 
	\\ \textit{The reformative push, initiated from Moscow in 1985, comes back like a boomerang towards the USSR}. \textbf{Lithuania, Latvia and Estonia} \underline{thus} \textbf{open themselves to the multiparty system.} \footnote{\label{note2} Free translation }  
\end{exe}


However, discourse connectives do not always mark the presence of discourse relations. For example, while the word \emph{`et'} is not a discourse connective in (\ref{ex:one}), it signals a \textsc{continuation} relation in (\ref{ex:two}).

\begin{exe}
    \ex \label{ex:two} \textit{La fédération CGT des transports s'est élevée contre ``l'absence de concertation''} \underline{et} \textbf{estime que les salariés ``n'ont rien de bon à attendre de cette restructuration''.}\textsuperscript{\ref{note1}}  
    \\ \textit{The CGT transport federation have risen against ``the lack of consultation''} \underline{and} \textbf{consider that employees have ``nothing positive to expect from this restructuring.''} \textsuperscript{\ref{note2}} 
    
\end{exe}

While studies have shown that discourse usage of discourse connectives can be accurately identified for English \cite{pitler09,lin14}, only a few studies have focused on the disambiguation of discourse connectives in other languages. In this paper, we investigate the usefulness of features proposed in the literature for the disambiguation of English discourse connectives for French discourse connectives. \footnote{The source codes along with the model trained for the disambiguation of French discourse connective are available at \url{https://github.com/mjlaali/french-dc-disambiguation}}

This paper is organized as follow: Section 2 reviews related work. Section 3 describes our approach to disambiguate French discourse connectives. Section 4 reports our results, and finally Section 5 presents our conclusions and future work.

\section{Related Work}
\label{sec:related-work}

With respect to discourse organization, discourse connectives constitute the most basic way of signaling the speaker’s or writer’s intentions. They provide an important clue to disambiguate discourse relations whose interpretations would be opaque without them \cite{asr12,drenhaus14,millis95,murray95,murray97}. Hence, lexicons of discourse connectives associated with the relations that they express can be very useful for discourse studies (e.g. developing discourse annotated corpora \cite{prasad08,danlos12,polakova13,al-saif10}, automatic discourse analysis \cite{xue15,lin14}, etc.) and have been developed for English \cite{knott96}, Spanish \cite{alonsoalemany02}, German \cite{stede98} and French \cite{danlos15}. 

Discourse connectives can be ambiguous at two levels: first, they can be used in \textit{discourse-usage} or \textit{non-discourse-usage}, and second, they may be used to signal more than one discourse relation. To automatically disambiguate discourse connectives, discourse annotated corpora such as the Penn Discourse Treebank (PDTB) \cite{prasad08} are instrumented. The PDTB is the largest corpus of discourse annotated texts. It contains articles from the Wall Street Journal, where discourse connectives that are used in \textit{discourse-usage} have been annotated by the discourse relation that they signal. The same approach has been used in the French Discourse Treebank (FDTB) \cite{danlos12}, however to date, only \textit{discourse-usage} and \textit{non-discourse-usage} of French discourse connectives have been annotated in the FDTB.

Most of previous work on the disambiguation of discourse connectives have focused on English discourse connectives \cite{marcu00,pitler09,lin14}. One of earliest and pioneer work on the disambiguation of discourse connectives, \citet{pitler09}, showed that four syntactic features (see Section \ref{sec:method:disam} for details about the features) and the connective itself can disambiguate discourse connectives with an accuracy of 95.04\% within the PDTB \cite{prasad08}. Their approach used these features not only to disambiguate discourse connectives between \emph{discourse-usage} and \emph{non-discourse-usage}, but also to tag the discourse relation signalled by the discourse connectives. 

Later, \citet{lin14} used the context of the connective (i.e. the previous and the following word of the connective) and added seven lexico-syntactic features to the feature set proposed by \citet{pitler09}. In doing so, \citeauthor{lin14} achieved an accuracy of 97.34\% for the disambiguation of discourse connectives in the PDTB.

On the other hand, the disambiguation of discourse connectives in languages other than English has received much less attention. Due to syntactic differences across languages and different discourse annotation methodologies, the techniques developed for one language may not be as effective in another. For example, English discourse connectives include mostly  subordinating conjunctions (e.g. \textit{when}) or coordinating conjunctions (e.g. \textit{but}). In addition, only a few connectives are disjoint (e.g. \textit{On the one hand ... On the other hand}). This is not the case for Chinese which uses many more disjoint connectives \cite{zhou12-a}. Inspired by \citet{pitler09}, \citet{alsaif11} proposed an approach for the disambiguation of Arabic Discourse connectives. \citeauthor{alsaif11} have shown that the features proposed by \citet{pitler09} work well for Arabic with an accuracy of 91.2\%. Moreover, they further improved the result of their system by considering Arabic-specific morphological features and achieved an accuracy of 92.4\%.

Today, due to the availability of discourse annotated corpora such as the French Discourse Treebank \cite{danlos15}, it is possible to analyse how the features developed for English behave when applied to French.

\section{Experiment}

\subsection{Corpus}

To evaluate the disambiguation of French discourse connectives, we used the French Discourse Treebank (FDTB) \cite{danlos15} which constitutes the largest publicly available discourse annotated corpus for French. The corpus contains the annotation of more than 10K connectives used in discourse usage in the French Treebank corpus \cite{abeille03}. The FDTB uses the French discourse connectives of the LEXCONN resource\footnote{The FDTB used the second version of the LEXCONN resource which is not publicly available yet.} \cite{roze12}, a lexicon of 328 French discourse connectives (e.g. \emph{`alors que'}) and their morphological variations (e.g. \emph{`alors qu' '}). Out of 328 discourse connectives listed in LEXCONN, 88 connectives appeared in the FDTB. 

For training and testing, we used the annotated discourse connectives of the FDTB as positive instances and all other occurrences of the connectives that were not annotated were used as negative instances. To compare the size of the FDTB dataset, we constructed a similar dataset from the PDTB. Table~\ref{tab:corpora-stat} shows the size of the datasets extracted from both the FDTB and the PDTB. 

\begin{table}[]
    \centering
    \caption{Statistics of the Datasets Extracted from the FDTB and the PDTB }
    \begin{tabular}{|l|r|r|r|}
\hline
             & \textbf{Positive Examples} & \textbf{Negative Examples} & \textbf{\# Words} \\ \hline
\textbf{FDTB} & 10K                       & 40K                        & 557K              \\ 
\textbf{PDTB} & 14K                       & 37K                        & 931K              \\ \hline
\end{tabular}
    \label{tab:corpora-stat}
\end{table}

As Table~\ref{tab:corpora-stat} shows, while the FDTB is smaller than the PDTB (10K instances of connectives in the FDTB vs. 14K instances in the PDTB), more types of discourse connectives are annotated in the FDTB (see Table~\ref{tab:distribution}). The FDTB contains the annotations of 372 different discourse connectives while the PDTB contains the annotations of 100 different discourse connectives. Table~\ref{tab:distribution} shows the distribution of the  discourse connectives in both corpora along with their frequency. Not surprisingly, 61\% (= 25\% + 36\%) of the French discourse connectives appear less than 10 times. This constitutes a large portion of French discourse connectives if we compare this number to its English counterpart in the PDTB (i.e. 18\%). This entails that it will be more difficult to learn an accurate model for the disambiguation of such low frequent discourse connectives.

\begin{table}[]
\centering
\caption{Distribution of Discourse Connectives (DCs) in the FDTB and the PDTB}
\begin{tabular}{|l|r|r|r|r|}
\hline
& \multicolumn{2}{|c|}{\textbf{FDTB (French)}} & \multicolumn{2}{|c|}{\textbf{PDTB (English)}} \\ \hline
\textbf{Frequency} & \textbf{Number of DCs} & \textbf{\%} & \textbf{Number of DCs} &  \textbf{\%} \\ \hline
\(f = 1 \) & 92 & 25\% & 3 & 3\%\\ \hline
\(1 < f < 10 \) & 133 & 36\% & 15 & 15\% \\ \hline
\(f \geq 10 \) & 147 & 39\% & 82 & 82\% \\ \hline
\textbf{Total} & \textbf{372} & \textbf{100\%} & \textbf{100} & \textbf{100\%} \\ \hline

\end{tabular}
\label{tab:distribution}
\end{table}

\subsection{Entropy of French Discourse Connectives}

To estimate the difficulty of the task for French compared to English, we compared the ambiguity of discourse connectives in the two languages by calculating the entropy of each discourse connective. Table~\ref{tab:stat} shows the top three most ambiguous and the top three least ambiguous discourse connectives (based on entropy) in the FDTB and the PDTB\footnote{To achieve statistically reliable results, we did not consider discourse connectives that appear less than 20 times.}. As Table~\ref{tab:stat} shows, the French discourse connectives \textit{`effectivement'} and \textit{`sinon'} are used as often in a discourse/non-discourse context (yielding an entropy of 1.0). On the other hand, in English \textit{`on the other hand'}, \textit{`particularly'} and \textit{`upon'} are the least ambiguous (entropy = 0.0) as they are always used to signal a discourse relation. 

\begin{table}[ht]
\centering
\caption{Entropy of Top Three Most/Least Ambiguous Discourse Connectives (DCs) in the FDTB and the PDTB}
    \begin{tabular}{|c|r|r|}
        \hline 
        \multicolumn{3}{|c|}{\textbf{FDTB (French)}} \\ \hline 
        {\textbf{DC}} &  {\textbf{Entropy}} & {\textbf{Frequency}}\\ \hline
        \textit{effectivement} & 1.00 & 27 \\ 
        \textit{sinon}         & 1.00 & 27 \\ 
        \textit{alors}         & 0.99 & 183\\ 
        \textit{...}           & ...  & ...\\
        \textit{toutefois} & 0.00   & 135  \\ 
        \textit{à}         & 0.00   & 9880 \\ 
        \textit{à propos}  & 0.00   & 35   \\ \hline
        \textbf{Average Entropy} &   \textbf{0.39} & \\ \hline
        
    \end{tabular}

\vspace*{1 em}

    \centering
    \begin{tabular}{|c|r|r|}
        \hline 
        \multicolumn{3}{|c|}{\textbf{PDTB (English)}} \\ \hline
        {\textbf{DC}} &  {\textbf{Entropy}} & {\textbf{Frequency}}  \\ \hline
        \textit{in contrast} & 1.00 & 22   \\ 
        \textit{as a result} & 1.00 & 135  \\ 
        \textit{besides}     & 1.00 & 32   \\ 
        \textit{...}         & ...  & ...  \\ 
        \textit{on the other hand} & 0.00   & 28  \\ 
        \textit{particularly}      & 0.00   & 130 \\ 
        \textit{upon}              & 0.00   & 41  \\ \hline
        \textbf{Average Entropy} &   \textbf{0.51} & \\ \hline
    \end{tabular}
\label{tab:stat}
\end{table}

Table~\ref{tab:stat} also shows the weighted average entropy of discourse connectives for each language. The entropy of French discourse connectives is 0.39 while the entropy of English discourse connectives is 0.51. This seems to indicate that the disambiguation of French discourse connectives can be considered a slightly easier task than the disambiguation of English discourse connectives.

To make a more detailed comparison, it would be preferable to align French and English discourse connectives with the same meaning and then compare the entropy of the mapped discourse connectives. Unfortunately, discourse connectives are language specific and cannot be easily aligned. For example, the accuracy of alignments achieved from statistical word-alignment models are very low for discourse connectives \cite{laali14}. To the best of our knowledge, a cross-lingual alignment of discourse connectives is available only for \textsc{casual} discourse connectives \cite{zufferey12}. \citet{zufferey12} manually aligned a few hundred occurrences of a discourse connective with their translation over the Europarl \cite{koehn05} parallel texts. Then, they created an English-French dictionary for discourse connectives based on the similarities and discrepancies between the discourse connectives and their most appropriate translation. 

\begin{table}[]
\centering
\caption{Entropy of Discourse Connectives (DCs) that signal a \textsc{Cause} relation in the FDTB and the PDTB}

\begin{tabular}[t]{|l|l|r|}
\hline
\textbf{French DC} & \textbf{English Translations}    & \textbf{Entropy} \\ \hline
\textit{parce que} & \textit{because}                 & 0.55             \\
\textit{puisque}   & \textit{since, as, because}      & 0.25             \\
\textit{car}       & \textit{because, as, since, for} & 0.05             \\ \hline
\end{tabular}

\vspace*{1 em}

\begin{tabular}[t]{|l|l|r|}
\hline
\textbf{English DC} & \textbf{French Translations}                               & \textbf{Entropy} \\ \hline
\textit{because}    & \textit{car, parce que}                                   & 0.98             \\   [.5em]
\textit{since}      & \parbox{0.25\textwidth}{\textit{puisque, étant donné que, car}}                    & 0.80             \\ [.5em]
\textit{as}         & \parbox{0.25\textwidth}{\textit{car, étant donné que, puisque, dans la mesure où}} & 0.59             \\ \hline
\end{tabular}

\label{tab:ent-casual}
\end{table}

Table~\ref{tab:ent-casual} shows the entropy of French and English discourse connectives  that signal the \textsc{Cause} relation with their most likely translations\footnote{Note that some translations of discourse connectives such as \textit{`étant donné que'} are not considered discourse connectives in the FDTB and the PDTB because they do not fit the formal definition of discourse connectives. Therefore, we do not list their entropy in Table~\ref{tab:ent-casual}.} as identified by \citet{zufferey12}. As Table~\ref{tab:ent-casual} shows, there does not seem to be a direct relationship between the entropy of the mapped discourse connectives. For example, while the French discourse connective \textit{`car'} has an entropy of 0.05 (i.e. \textit{`car'} is more than 99\% of the time used in \textit{discourse-usage} in the FDTB), its translations in English (i.e.  \textit{`because'}, \textit{`since'}, and \textit{`as'}) are very ambiguous.

The disparity between the entropy of discourse connectives in the FDTB and the PDTB can be explained by the differences between the languages and the annotation methodology. Regardless of its source, this disparity motivated us to investigate the applicability of features proposed for the disambiguation of English discourse connectives for French.



\subsection{Features}

As mentioned in Section~\ref{sec:related-work}, \citet{pitler09} have shown that the context of discourse connectives in the syntactic tree is very discriminating for the disambiguation of English discourse connectives. They proposed four syntactic features: 
\begin{enumerate}
\item \emph{SelfCat}: The highest node in the parse tree that covers the connective words but nothing more.
\item \emph{SelfCatParent}: The parent of the \emph{SelfCat}.
\item \emph{SelfCatLeftSibling}: The left sibling of the \emph{SelfCat}.
\item \emph{SelfCatRightSibling}: The right sibling of the \emph{SelfCat}.
\end{enumerate}

%

\begin{figure*}[t]
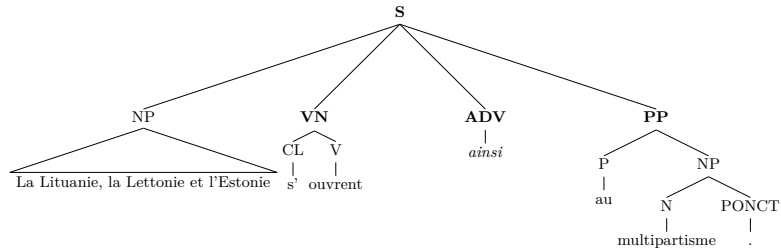

\centering
\resizebox{\linewidth}{!}{ 
     \Tree [ .\textbf{S} \qroof{La Lituanie, la Lettonie et l'Estonie}.NP [ .\textbf{VN} [ .CL s' ] [ .V ouvrent ] ] [ .\textbf{ADV} \emph{ainsi} ] [ .\textbf{PP} [ .P au ] [ .NP [ .N multipartisme ] [ .PONCT . ] ] ] ] 
}

\caption{The Parse Tree\textsuperscript{5} of the Second Sentence of Example (\ref{ex:one})
}
\label{fig:example}
\end{figure*}

\footnotetext{The parse tree is taken from the French Treebank.}

To illustrate these four features, consider the parse tree of the second sentence in Example (\ref{ex:one}) shown in Figure~\ref{fig:example} and the discourse connective \emph{`ainsi'}. The \emph{SelfCat} node is the \emph{`ADV'} node in the parse tree and its parent, left and right siblings are the \emph{`S'}, \emph{`VN'} and \emph{`PP'} nodes, respectively.

In addition to the four features above, \citet{pitler09} used the discourse connective itself (case sensitive) as an additional feature for the classifier. The purpose of using the case sensitive version is to distinguish connectives positioned at the beginning of sentences. We slightly modified this feature by using the case-folded version of the discourse connective (called the \emph{Conn} Feature). However, we created a new feature (called the \emph{Pos} Feature) to explicitly indicate the position of the discourse connective within the sentence (i.e. \emph{at-the-beginning} or \emph{not-at-the-beginning}). These two features are as informative as the case-sensitive connective string proposed by \citet{pitler09}, however, separating these features gives the classifier more flexibility when building its model. In Example (\ref{ex:one}), these two features are \emph{`ainsi'} and \emph{`not-at-the-beginning'}, respectively.

\subsection{Data Preparation}
\label{sec:method:disam}

Although the focus of the work is the disambiguation of French discourse connectives, we performed the same experiments with English discourse connectives as well. This allowed us to better analyse the results and make a comparison between the performance of our model for French and for English. For our experiments, we used the FDTB and the PDTB corpora for gold discourse annotation for French and English texts respectively. We also used the French Treebank (FTB)~\cite{abeille03} and Penn Treebank (PTB)~\cite{marcus93} to obtain the parse trees of the French and English texts.

To prepare the FDTB for our experiments, we used the French Treebank (FTB)~\cite{abeille03} to obtain the syntactic trees of the FDTB sentences. Next, we converted the FDTB sentences along  with their syntactic trees into the CoNLL-2015 shared task format \cite{xue15}. The English version of our experiments were performed on the CoNLL 2015 shared task dataset  \cite{xue15} which contains the annotations of the PDTB and parse trees of the PTB. Similarly to \citet{pitler09-a}, we used sections 2-22 of the PDTB for our experiments. 

To annotate discourse connectives, the input sentences were first searched for terms that match a discourse connective. Then, inspired by \citet{pitler09-a}, a binary classifier was built using six local syntactic and lexical features to classify discourse connectives as \emph{in-discourse-usage} or \emph{not-in-discourse-usage}.

\section{Results and Analysis}

\subsection{Results}
Similarly to \citet{pitler09}, we report results using a maximum entropy classifier using ten-fold cross-validation over the extracted datasets. We used the off-the-shelf implementation of the maximum entropy classifier available in WEKA \cite{hall09} for our experiments. 

Table~\ref{tab:overall-performance} shows the overall performance of the classifier for the disambiguation of French and English discourse connectives. The results show that the classifier can distinguish between \emph{discourse-usage} and \emph{non-discourse-usage} of French discourse connectives with an accuracy of 94.2\% and an F-Measure of 86.2\%. This is close to the results achieved for English discourse connectives over the PDTB (accuracy of 93.6\% and F-score of 88.9\%). 

\begin{table}[]
\centering
\caption{Overall Performance of the Disambiguation of Discourse Connectives}
\label{tab:overall-performance}
\begin{tabular}{|p{3.5cm}|r|r|r|r|}
\hline
\textbf{Dataset}                 & \textbf{Precision} & \textbf{Recall} & \textbf{F-Measure} & \textbf{Accuracy} \\ \hline
\textbf{Extracted from the FDTB} & 85.7\%               & 86.7\%            & 86.2\%               & 94.2\%              \\ 
\textbf{Extracted from the PDTB} & 87.1\%               & 90.8\%            & 88.9\%               & 93.6\%              \\ \hline
\end{tabular}
\end{table}

\subsection{Feature Analysis}
To evaluate the contribution of each feature, we ranked the features by their information gain  for both languages. As Table~\ref{tab:featuresInfoGain} shows, with our datasets, the syntactic features provide less information about \textit{discourse-usage} or \textit{non-discourse-usage} for French discourse connectives than they do for English. For example, the \textit{Selfcat} feature has a significantly lower information gain than the \textit{Conn} feature for the disambiguation of French discourse connectives while this is not the case for English discourse connectives. This seems to indicate that the English discourse connectives tend to appear in more restricted syntactic contexts than French discourse connectives.

\begin{table*}[]
\centering
\caption{Information Gain of Each Feature in the Disambiguation of Discourse Connectives }
\begin{tabular}{|l|l|r|r|}
\cline{2-4}
\multicolumn{1}{c|}{} & \textbf{Feature}    & \textbf{French} & \textbf{English} \\ \hline
\textit{Lexical:} & \textit{Conn}            & 0.352           & 0.351            \\ \hline
\textit{Syntactic:} & \textit{SelfCat}             & 0.167           & 0.468            \\ 
& \textit{SelfCatLeftSibling}  & 0.108           & 0.145            \\ 
& \textit{SelfCatParent}       & 0.093           & 0.292            \\ 
& \textit{Pos}             & 0.045           & 0.119            \\ 
& \textit{SelfCatRightSibling} & 0.032           & 0.085            \\ \hline
\end{tabular}
\label{tab:featuresInfoGain}
\end{table*}

We also experimented with different features combinations. We ranked the features by their information gain and added one feature at a time.  Table~\ref{tab:featuresSetPerformance} shows the accuracy of the classifier for each subset of features for French discourse connectives. The differences between the accuracies were evaluated with the Student t test, with P~<~0.05. Statistically significant increases are marked with $\Uparrow$ in the table while a lack of statistically significant increase is indicated with a $\oslash$. As Table~\ref{tab:featuresSetPerformance} shows, using only the connective text (the \textit{Conn} feature) gives an overall accuracy of 89.10\%, which is a reasonably high baseline. Adding the \textit{SelfCat}, \textit{SelfCatLeftSibling} and \textit{SelfCatParent} features gradually improves the accuracy from 89.10\% to 94.21\%. Table~\ref{tab:featuresSetPerformance} shows that the effects of the \textit{Pos} and \textit{SelfCatRightSibling} features are negligible and without these two features the accuracy of the classifier (i.e. 93.21\%) does not yield a statistically lower accuracy than the overall result (94.52\%) which combines all features. 

\begin{table}[]
\centering
\caption{The Accuracy of the Classifier for Each Feature Set}
\label{tab:featuresSetPerformance}
\begin{tabular}{|p{2.5in}|P{4em}p{1em}|P{4em}p{1em}|}
\hline
\textbf{Feature Set}                                                                                    & \multicolumn{2}{|p{5em}|}{\textbf{French (FDTB)}} & \multicolumn{2}{|p{5em}|}{\textbf{English (PDTB)}} \\ \hline
\textit{Conn}                                                                                           & 89.10\%       &            & 85.38\%        & \\ \hline
\textit{Conn + SelfCat}                                                                                 & 91.02\%       & $\Uparrow$ & 93.15\%        & $\Uparrow$ \\ \hline
\textit{Conn + SelfCat + SelfCatLeftSibling}                                                            & 91.77\%       & $\Uparrow$ & 93.38\%        & $\Uparrow$\\ \hline
\textit{Conn + SelfCat + SelfCatLeftSibling + SelfCatParent}                                            & 94.21\%       & $\Uparrow$ & 93.42\%        & $\oslash$ \\ \hline
\textit{Conn + SelfCat + SelfCatLeftSibling + SelfCatParent + Pos }                                     & 94.16\%       & $\oslash$  & 93.58\%        & $\Uparrow$\\ \hline
\textit{Conn + SelfCat + SelfCatLeftSibling + SelfCatParent + Pos + SelfCatRightSibling}                & 94.23\%       & $\oslash$  & 93.52\%        & $\oslash$ \\ \hline
\end{tabular}
\end{table}

As Table~\ref{tab:featuresSetPerformance} also shows, the effect of each feature is different for English. Using only the \textit{Conn} feature gives an accuracy of 85.38\% which is lower than the accuracy achieved for French (i.e. 89.10\%). Adding the \textit{SelfCat} feature improves the accuracy from 85.38\% to 93.15\%. The effect of the rest of the features is negligible and only improves the accuracy to 93.52\%. This again seems to show that English discourse connectives tend to appear in more restricted syntactic contexts than French discourse connectives.

\subsection{Per-Connective Analysis}

The overall results of Table~\ref{tab:overall-performance} showed that the features proposed for English can also accurately disambiguate French discourse connectives. However, if we analyse the results for each connective, many seem to be very well classified with the features used; while a few are more difficult to disambiguate. If we use as a baseline the assignment of the most likely class based only on the \textit{Conn} feature, many connectives obtained statistically significant improvements with the classifier learned over the proposed features. Table~\ref{tab:res:best} shows the accuracy of the classifier for the French discourse connectives which achieved the greatest improvements over the baseline. Again, the differences between the accuracies were evaluated with the Student t test, with P~<~.05 considered statistically significant and marked with $\Uparrow$ and lack of statistical increase is indicated by $\oslash$ in the table. As Table~\ref{tab:res:best} shows, for these connectives, the classifier can disambiguate \textit{discourse-usage} versus \textit{non-discourse-usage} with a much better accuracy than the baseline. For example, the classifier can disambiguate \textit{`sinon'}, which is among the top tree ambiguous French discourse connectives (see Table~\ref{tab:stat}), with an accuracy of 88.89\%, yet only 27 instances of this connective are available in the dataset.

\begin{table}[ht]
\centering
\caption{Accuracy of the Classifier for the French Discourse Connectives (DCs) that Achieved the Greatest Improvement over the Baseline}
\label{tab:res:best}
\begin{tabular}{|l|rr|rrrl|}
\hline
\textbf{DC}         & \multicolumn{1}{l}{\textbf{Freq.}} & \multicolumn{1}{l|}{\textbf{Entropy}} & \multicolumn{1}{l}{\textbf{Baseline}} & \multicolumn{1}{l}{\textbf{Accuracy}} & \multicolumn{1}{l}{\textbf{Diff.}} & \\ \hline
\textit{sinon}      & 27            & 1.00             & 33.33\%           & 88.89\%           & 55.56\%   & $\Uparrow$     \\
\textit{aussi}      & 533           & 0.97             & 59.29\%           & 91.56\%           & 32.27\%    & $\Uparrow$    \\
\textit{au lieu de} & 37            & 0.88             & 70.27\%           & 97.30\%           & 27.03\%    & $\Uparrow$    \\
\textit{et}         & 8662          & 0.81             & 74.82\%           & 90.88\%           & 16.06\%   & $\Uparrow$ \\ \hline
\end{tabular}
\end{table}

While the accuracy of the classifier is high for many discourse connectives, there are a few discourse connectives that the classifier cannot disambiguate. The  five  discourse  connectives\footnote{To achieve statistically reliable results, we did not consider discourse connectives that appear less than 20 times.} that  achieve  the  lowest  accuracy  are listed in Table~\ref{tab:res:worst}. All the discourse connectives in Table~\ref{tab:res:worst} have very high entropy. For example, both `\textit{effectivement}' and `\textit{alors}' are among the top three ambiguous discourse connectives (see Table~\ref{tab:stat}). Even though the accuracy of the classifier is higher than the baseline (except for the \textit{`maintenant'} discourse connective), the increase is small or not statistically significant. For example, the accuracy for the discourse connective `\textit{effectivement}' is 55.56\% which is not statistically better than the baseline. These results show that for some connectives, the features proposed for English are sufficient (see Table~\ref{tab:res:best}), but for others, using only the connective and the syntactic features is not sufficient.

\begin{table}[]
\centering
\caption{Accuracy of the Classifier for Discourse Connectives With the Least Accuracy}
\label{tab:res:worst}
\begin{tabular}{|l|rr|rrrl|}
\hline
\textbf{DC}         & \multicolumn{1}{l}{\textbf{Freq.}} & \multicolumn{1}{l|}{\textbf{Entropy}} & \multicolumn{1}{l}{\textbf{Baseline}} & \multicolumn{1}{l}{\textbf{Accuracy}} & \multicolumn{1}{l}{\textbf{Diff.}} & \\ \hline
\textit{effectivement} & 27            & 1.00             & 25.93\%           & 55.56\%           & 29.63\%          & $\oslash$              \\
\textit{alors}         & 183           & 0.99             & 54.64\%           & 67.21\%           & 12.57\%         & $\Uparrow$               \\
\textit{auparavant}    & 21            & 0.99             & 57.14\%           & 61.90\%           & 4.76\%           & $\oslash$              \\
\textit{de même}       & 52            & 0.99             & 55.77\%           & 69.23\%           & 13.46\%           & $\Uparrow$  \\ 
\textit{maintenant}	& 81	& 0.93	& 65.43\%	& 62.96\%	& -2.47\%	& $\oslash$
\\ \hline 
\end{tabular}
\end{table}

\section{Conclusion and Future Work}
In this paper, we have investigated the applicability of the syntactic and lexical features proposed by \citet{pitler09} for the disambiguation of English discourse connectives for French. Our experiments on the French Discourse Treebank (FDTB) show that even though the syntactic features are less informative for the disambiguation of French discourse connectives than for English discourse connectives, overall the features can effectively disambiguate French discourse connectives between \emph{discourse-usage} and \emph{non-discourse-usage} as well in French as in English. The fact that the local syntactic features proposed for English can be used almost as effectively for French and Arabic \cite{alsaif11} suggests that lexicalized discourse connectives share certain common structural features cross-linguistically and that these structures are potentially an important component in discourse processing. However, our analysis also shows that the features are not as effective for all connectives. Some high entropy connectives such as \textit{`sinon'} have a very high accuracy whereas others such as `\textit{effectivement}' or \textit{`maintenant'} require additional features.

As future work, we would like to investigate features that do not need parse trees (such as the features proposed by Lin \textit{et al.} \cite{lin14}) for the disambiguation of discourse connectives. We believe that such features would be useful for languages that lack robust syntactic parsers.

\section*{Acknowledgments}
The authors would like to thank Université Paris-7 (Laboratoire de linguistique formelle) for giving us access to the FDTB and the FTB. The authors would also like to thank the anonymous reviewers for their feedback on the paper. This work was financially supported by NSERC.

\end{document}